\documentclass[10pt,twocolumn,letterpaper]{article}

\usepackage{iccv}              

\usepackage{tabularx}
\usepackage{rotating}
\usepackage{subcaption}

\newcommand{\no}{{\ding{55}}}
\newcommand{\yes}{{\ding{51}}}
\definecolor{mygray}{gray}{0.94}
\usepackage{pifont}
\usepackage{multirow}
\usepackage{colortbl}
\definecolor{iccvblue}{rgb}{0.21,0.49,0.74}
\usepackage[pagebackref,breaklinks,colorlinks,allcolors=iccvblue]{hyperref}

\usepackage{algorithm}
\usepackage{algpseudocode}
\usepackage{cuted}

\def\paperID{4129} 
\def\confName{ICCV}
\def\confYear{2025}

\title{Learning Efficient and Generalizable Human Representation \\ with Human Gaussian Model}  

\author{Yifan Liu$^{1,*}$, Shengjun Zhang$^{1,*}$, Chensheng Dai$^{1}$, Yang Chen$^{3}$, Hao Liu$^{2}$, Chen Li$^{2}$, Yueqi Duan$^{1\dag}$\\
$^{1}$Tsinghua University, $^{2}$WeChat Vision, Tecent Inc., $^{3}$Nanyang Technological University \\
{\tt\small \{liuyifan22, zhangsj23\}@mails.tsinghua.edu.cn, duanyueqi@tsinghua.edu.cn}}

\begin{document}
\maketitle

\begin{strip}
    \vspace{-1.8cm}
    \begin{center}
    \textbf{\url{https://github.com/Simon-Dcs/Human_Gaussian_Graph/}}
    \end{center}
    \centering
    \includegraphics[width=\linewidth]{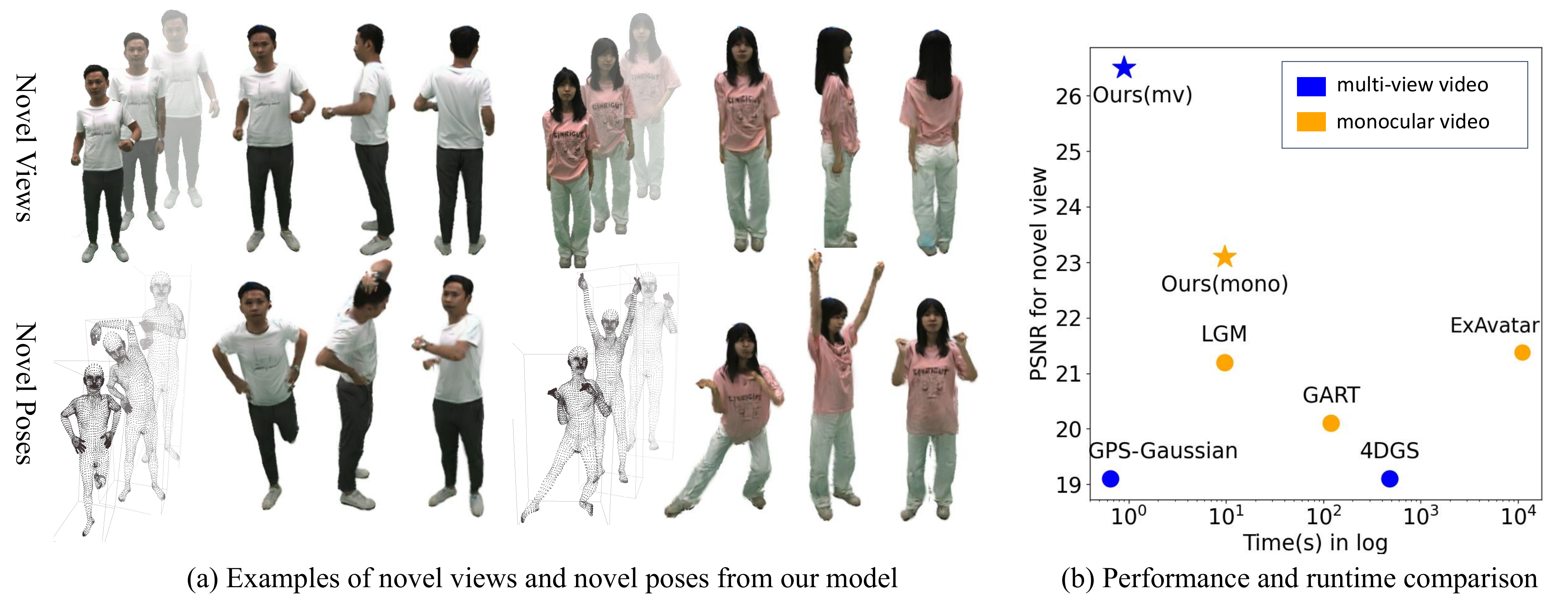}
    \vspace{-0.4cm}
    \captionof{figure}{Our method achieves state-of-the-art rendering quality while maintaining remarkable fast run-time performance. (a) Qualitative results: Our approach delivers high-fidelity results for both novel view synthesis and novel pose animation. (b) Performance comparison: Our method achieves the highest PSNR in both single view (yellow) and multi-view (blue) settings with superior computational efficiency.}
    \label{fig:teaser}
\end{strip}




\begin{abstract}

Modeling animatable human avatars from videos is a long-standing and challenging problem.
While conventional methods require per-instance optimization, recent feed-forward methods have been proposed to generate 3D Gaussians with a learnable network.
However, these methods predict Gaussians for each frame independently, without fully capturing the relations of Gaussians from different timestamps.
To address this, we propose Human Gaussian Graph to model the connection between predicted Gaussians and human SMPL mesh, so that we can leverage information from all frames to recover an animatable human representation.
Specifically, the Human Gaussian Graph contains dual layers where Gaussians are the first layer nodes and mesh vertices serve as the second layer nodes.
Based on this structure, we further propose the intra-node operation to aggregate various Gaussians connected to one mesh vertex, and inter-node operation to support message passing among mesh node neighbors. Experimental results on novel view synthesis and novel pose animation demonstrate the efficiency and generalization of our method.

\end{abstract}    
\section{Introduction}
\label{sec:intro}
Realistic human reconstruction is a fundamental task in computer vision with widespread application in virtual reality, gaming, healthcare and social media.
Remarkable progress has been made using neural implicit representations~\cite{Occupancy:2019, park2019deepsdf, mildenhall2020nerf} to model flexible topology, but these methods~\cite{PIFu:2019, PIFuHD:2020, alldieck2022photorealistic, SHERF:2023, ELICIT:2023} suffer from expensive time consumption in training and rendering.
Recently, 3D Gaussian Splatting~\cite{kerbl3Dgaussians} (3DGS) has drawn increasing attention for explicit Gaussian representations and real-time rendering performance. 
Benefiting from rasterization-based rendering, 3DGS avoids dense points querying in scene space so that it can maintain high efficiency and quality.

Since conventional methods~\cite{lei2023gart, qian20233dgsavatar, hu2023gaussianavatar, qian2024gaussianavatars, HAHA, GVA, SplattingAvatar} based on 3DGS require per-instance optimization, recent human reconstruction studies~\cite{zheng2024gpsgaussian, pan2024humansplat, LGM:2024} focus on directly regress Gaussian parameters with feed-forward networks.
Typically, these methods generate Gaussians with U-Net architectures~\cite{zheng2024gpsgaussian} or latent reconstruction transformers~\cite{pan2024humansplat} for each frame independently, without leveraging complementary information from other frames.
Besides, the lack of alignment between Gaussian representations and the human SMPL~\cite{SMPL:2015, SMPL-X:2019} mesh further limits the application of novel pose animation in downstream tasks.

To tackle these challenges, we propose Human Gaussian Graph to construct the relations of Gaussian groups from multiple frames and the human structure priors from SMPL mesh. 
Specifically, our Human Gaussian Graph contains dual layers, where Gaussians from all frames are the first-layer nodes, and the SMPL vertices, which are equivalent throughout the temporal axis, are the second-layer nodes.
Then, we define the edges between two-layer nodes as the alignment of Gaussians and SMPL vertices based on the spatial relations.
We present intra-node operations between two-layer nodes to aggregate temporal information from multiple frames.
Since the second-layer nodes are naturally connected according to human structures, we further introduce inter-node operation to support message passing between connected SMPL vertices in local regions.
As shown in Figure \ref{fig:teaser}, our method can generate high-quality generalizable and animatable human gaussian representations from videos.
We conduct experiments on novel view synthesis and novel pose animation. While the optimization-based methods~\cite{Wu_2024_CVPR, lei2023gart, moon2024exavatar} require expansive training time, our method is more efficient in a feed-forward manner. While other generalizable methods fail to benefit from the multiple frames of videos, our graph structure models the relations of temporal information to generate mesh-aligned animatable human Gaussians. 
As shown in Figure \ref{fig:teaser}, we surpass previous methods with less time consumption.
Our main contributions are summarized as follows:
\begin{itemize}
    \item We introduce Human Gaussian Graph to effectively model the relations of cross-frame Gaussian primitives and human SMPL mesh in videos.
    \item We present intra-node operations and inter-node operations to process the Human Gaussian Graph, enabling the interaction and aggregation across different nodes.
    \item Experimental results illustrate that our method can generate high-quality generalizable and animatable human Gaussian representations from videos.
\end{itemize}


\section{Related Work}
\noindent\textbf{3D Human Reconstruction.}
Reconstructing 3D humans is a research focus. The parametric template models~\cite{smpl,smplx, flame, mano} regulate a strong geometric prior of the human body, fueling a surge of research on human body poses~\cite{kocabas2020vibe, sun2021monocular, sun2023trace, goel2023humans, wang2023refit}. However, the explicit and predefined topology of parametric meshes cannot capture personalized appearances such as hair, and clothing~\cite{peopelsnap, alldieck2018video, habermann2019livecap}. Implicit methods~\cite{saito2019pifu,alldieck2022photorealistic,han2023high, HumanSGD:2023, HumanLRM:2024, he2020geoPifu, zhang2023globalcorrelated, Jiang_2023_CVPR}, utilizing SDF and NeRF, enables accurate depiction of 3D clothed humans. Methods like GTA~\cite{zhang2023globalcorrelated} use transformers to map input images into 3D tri-plane features, which are then used for reconstruction. Recently, 3D Gaussian Splatting~\cite{kerbl3Dgaussians} has become the new trend in human reconstruction~\cite{lei2023gart, jena2023splatarmor, qian20233dgsavatar, hu2023gaussianavatar, qian2024gaussianavatars, HAHA,GVA, SplattingAvatar}. Optimization-based methods such as GART ~\cite{lei2023gart} bind Gaussians onto the SMPL model and utilizes LBS skinning to map the Gaussians to poses in respective frames for supervision. Though high in quality, these methods are neither instant nor generalizable due to the time-consuming optimization process, limiting their downstream applications. We propose a method that effectively bypasses the optimization process and constructs high-quality generalizable avatars within inference time.

\noindent\textbf{Generalizable Gaussian Model.}
 pixelSplat~\cite{charatan2023pixelsplat} and MVSplat~\cite{chen2024mvsplat} are representative works showing that 3D Gaussians can be directly predicted from image pairs via feed-forward models, avoiding the time-consuming optimizing process. Methods like LGM~\cite{LGM:2024,GRM:2024, GSLRM:24, zhang2024GGN} predict Gaussian attributes for each input pixel in each view using large deterministic models with scaled training, and combine them as the final output. GPS-Gaussian~\cite{zheng2024gpsgaussian} focuses on novel view synthesis by splatting each pixel into space based on the estimated depth, showcasing in experiments that the generalizable Gaussian models have potential in human reconstruction. The most relevant work to ours is HumanSplat~\cite{pan2024humansplat}, which introduces human geometric priors into feed-forward Gaussian networks. HumanSplat tokenizes the SMPL~\cite{smpl,smplx} mesh and lets the image features attend to the SMPL tokens, injecting geometric priors into the pipeline. Though effective, their method is unable to deal with video inputs and cannot produce animatable avatars, In contrast, our method directly builds a graph based on the geometric priors to integrate cross-frame information, and our model outputs SMPL-aligned pose-driven Gaussians, opening up new possibilities for downstream applications. To our knowledge, we are the first work to achieve animatable Gaussian avatars within inference time.
\section{Method}
\begin{figure*}[t]
\centering
\includegraphics[width=1.0\linewidth]{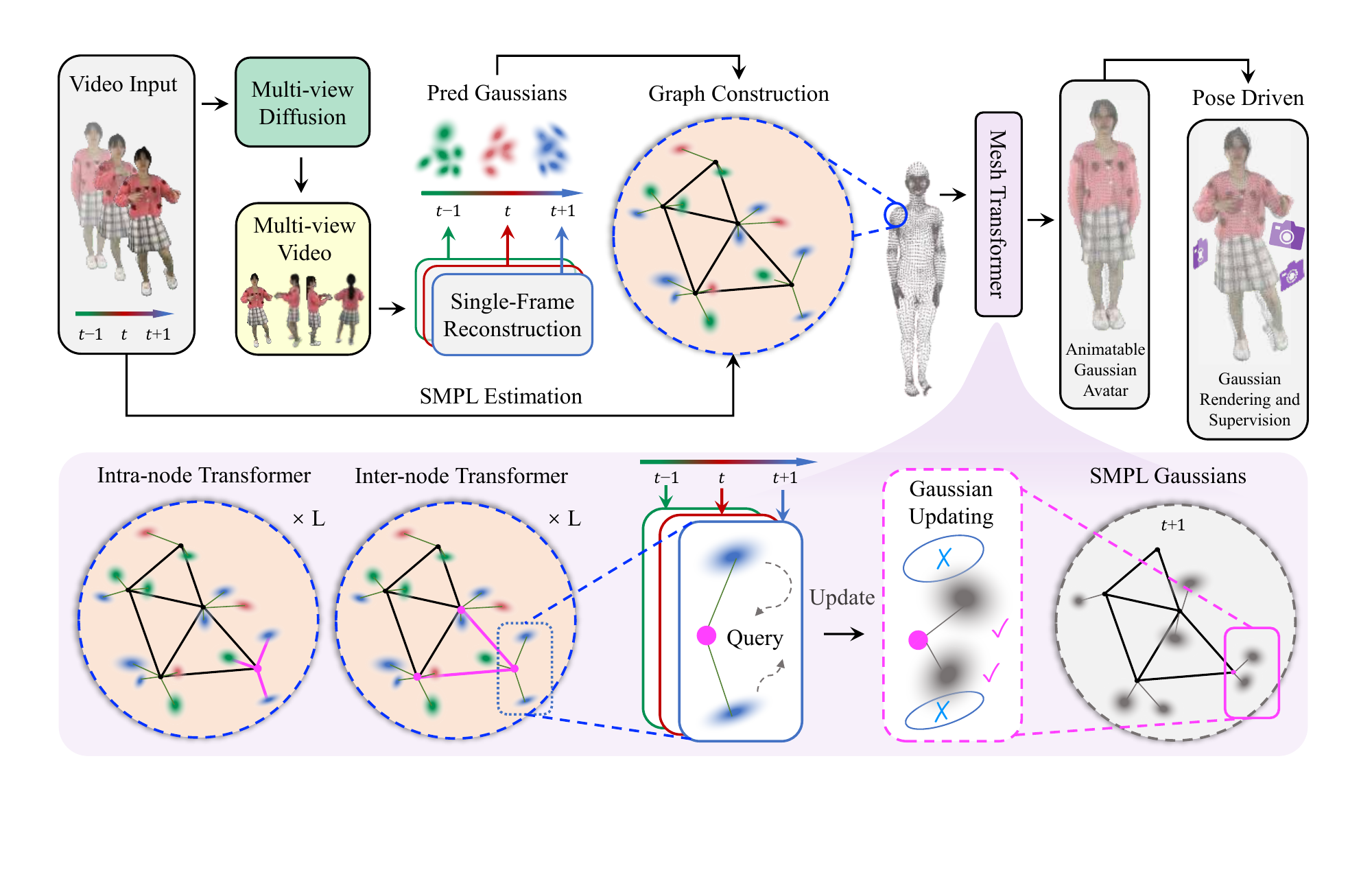}
\vspace{-2cm}
\caption{Overview. Given an input human video, our goal is to build high-fidelity animatable Gaussian representations within inference time. We first establish frame-wise Gaussian representations through a feed-forward 3DGS network. Then we construct Human Gaussian Graph (HGG) to model the relations between predicted Gaussians from multiple frames and the SMPL mesh (Section~\ref{meshgraph}). We introduce two complementary types of operations on the HGG: the intra-node operation that extracts temporal features across multiple timesteps, and the inter-node operation that facilitates robust local message passing between topologically adjacent nodes (Section~\ref{inter}). Finally, the Gaussians are updated into SMPL-aligned Gaussians through the HGG framework, enabling novel pose animation. (Section~\ref{smpl-aligned}) }
\vspace{-0.3cm}
\label{fig:pipeline}
\end{figure*}

The overall framework is illustrated in Figure~\ref{fig:pipeline}. Given a video $\{I^{t}\}_{t=1}^{T}$, our goal is to reconstruct an animatable avatar within inference time. In our framework, we first build Gaussian representations $G^{t}=\{\mu^{t}, f^{t}\}$ for each frame $I^{t}$ through a feed-forward 3DGS network~\cite{LGM:2024} with generative priors~\cite{long2023wonder3d}. 
Then, we construct Human Gaussian Graph to model the relations between predicted Gaussians from multiple frames and the SMPL mesh~\cite{SMPL:2015}, where Gaussians are the first layer nodes and SMPL vertices serve as the second layer nodes.
To enable cross-frame Gaussian aggregation, we introduce an intra-node operation to extract features from different timesteps on each SMPL vertices.
Furthermore, we design an intra-node operation to support message passing between SMPL vertices and their neighbors. 
In this way, our model reconstructs SMPL-aligned Gaussians that can be driven by pose, fueling downstream application like virtual reality and video games.

\subsection{Preliminaries}

\noindent\textbf{SMPL model.} SMPL~\cite{smpl} is a parametric human mesh model, which is created by skinning and blend shapes. The shape parameters $\beta\in\mathbb{R}^{10}$ adjust the body shape and the pose parameters $\theta\in\mathbb{R}^{24\times3}$ re-poses the body to various gestures with Linear Blending Skinning (LBS). A 3D point $x$ in the canonical space is transformed to a posed space defined by $\theta$ as:
\begin{equation}
    \tilde{x}=\sum_{k=1}^{K}w_{k}\left(G_{k}(J,\theta)x+b_{k}(J,\theta,\beta)\right),
    \label{eq:LBS}
\end{equation}
where $J$ includes $K$ joint locations, $G_{k}(J,\theta)$ and $b_{k}(J,\theta,\beta)$ are the transformation matrix and translation vector of joint $k$, and $w_{k}$ is the linear blend weight.

\noindent\textbf{3D Gaussian Splatting.} 
3D Gaussian Splatting~\cite{kerbl20233d} represents 3D objects or scenes using a set of Gaussians, including a center position $\mu\in\mathbb{R}^3$, an opacity $\alpha\in\mathbb{R}$, a covariance matrix $\Sigma\in\mathbb{R}^{3\times 3}$ and spherical harmonics denoting the color $\mathbf{c}\in\mathbb{R}^k$. 
The Gaussian function can be formulated as:
\begin{equation}
    G(x)=e^{-\frac{1}{2}(x-\mu)^{\top}\Sigma^{-1}(x-\mu)},
\end{equation}
where $\Sigma=RSS^\top R^\top$, $S$ is the scaling matrix and $R$ is the rotation matrix.
For every pixel, the color is rendered by a set of Gaussians sorted in depth order:
\begin{equation}
    C=\sum_{i\in N} c_{i}\alpha_{i}\prod_{j=1}^{i-1}(1-\alpha_{i}). \label{eq: color rendering}
\end{equation}

\subsection{Human Gaussian Graph} \label{meshgraph}

\begin{algorithm}[t]
\caption{Constructing the Graph}\label{alg:construct}
\renewcommand{\algorithmicrequire}{\textbf{Input:}}
\renewcommand{\algorithmicensure}{\textbf{Output:}}
\begin{algorithmic}
\Require Input video $\{I^t\}_{t=1}^T$, SMPL poses $\{p^{t}\}_{i=1}^{N}$, 
\Ensure Human Gaussian Graph $\{\mathcal{V},\mathcal{G},\mathcal{E}_{vv},\mathcal{E}_{vg}\}$

\For {$t \gets 1$ \textbf{to} $T$}
    \State $G^{t}=\{g_m^t\}_{m=1}^M=\Phi_{\text{G}}(I^{t})$ \Comment per-frame Gaussians
\EndFor

\For {$t \gets 1$ \textbf{to} $T$}
    \State $\mathcal{V}^{t}=\{v_{i}^{t}\}_{i=1}^{N} = \psi_{\text{LBS}}(\mathcal{V},p^t)$
    \For{$m \gets 1$ \textbf{to} $M$}
        \State $k = \mathop{\arg\min}\limits_{i}\ \psi_{\text{d}}(\mu_{m}^t, v_i^t)$
        \State $\mathcal{E}_{vg}(g_m^t,v_{k})=1$ \Comment{Gaussian-Mesh edges}
    \EndFor
\EndFor
\For {$s \gets 1$ \textbf{to} $N$}
    \For {$s \gets 1$ \textbf{to} $N$}
        \If {$d(v_{s},v_{n})\leq d_0 $} 
            \State $\mathcal{E}_{vv}(v_{s},v_{t})= 1$ \Comment Mesh-level edges
        \EndIf
    \EndFor
\EndFor
\end{algorithmic}
\end{algorithm}

\noindent\textbf{Dual-layer Nodes}
Given the input video $\{I^{t}\}_{t=1}^{T} \in \mathbb{R}^{T\times H\times W\times3}$, we generate multi-view images with a diffusion model~\cite{long2023wonder3d}, and predict the corresponding set of Gaussian for each frame $I^t$:
\begin{equation}
    G^{t}=\{g_m^t\}_{m=1}^M=\Phi_{\text{G}}(I^{t}), \quad g_m^t=\{\mu_{m}^t, f_{m}^t\},
\end{equation}
where $\Phi_{\text{G}}$ denotes the Gaussian prediction model, $\mu_{m}^t\in\mathbb{R}^{3}$ is the Gaussian center, and $f_{m}^t\in\mathbb{R}^{C}$ is the Gaussian features.
We define $\mathcal{G}=\{G^{t}\}_{t=1}^{T}$ as the first-layer nodes, serving as the source of the information needed for reconstruction. The SMPL vertices $\mathcal{V}=\{v_{i}\}_{i=1}^{N}$ serve as the second-layer nodes, which will be leveraged as a set of equivalent points throughout the temporal axis. 

\noindent\textbf{Gaussian-Mesh Edges.} 
To model the relations between Gaussians in different frames, we build the edges $\mathcal{E}_{vg}\in\{0,1\}^{|\mathcal{G}|\times |\mathcal{V}|}$ between Gaussian nodes $\mathcal{G}$ and mesh vertices $\mathcal{V}$.
Given a frame $I^t$, we fisrt estimate the SMPL poses $p^t=\{\beta^{t}, \theta^{t}\}$ for the current motion of the human. 
We transform the second-layer nodes $\mathcal{V}$ into the pose of current frame:
\begin{equation}
    \mathcal{V}^{t}=\{v_{i}^{t}\}_{i=1}^{N} = \psi_{\text{LBS}}(\mathcal{V},p^t),
\end{equation}
where $\psi_{\text{LBS}}$ denotes the LBS algorithm in Eq.~\ref{eq:LBS}.
For each Gaussian $g_m^t \in \mathcal{G}^t$, we connect it to its closet node $v_{k}^{t}$ in the second layer:
\begin{equation}
    \mathcal{E}_{vg}(g_m^t,v_{k})=1, \quad \text{where}\ k = \mathop{\arg\min}\limits_{i}\ \psi_{\text{d}}(\mu_{m}^t, v_i^t),
\end{equation}
where $\psi_{\text{d}}$ stands for the Euclidean distance.
Notably, Gaussians from different frames are connected to the same set of second-level nodes $\mathcal{V}$, thus reorganizing the Gaussians from temporal grouping to spatial grouping. 

\noindent\textbf{Mesh-level Edges.} 
The SMPL mesh provides a natural connectivity for SMPL vertices. 
We connect vertices with its neigbours in the SMPL mesh. Specifically, let $V_n$ be a second-layer vertex. 
We define a distance $d(n,n')$ on the mesh, denoting the number of mesh faces needed to connect the two vertices. 
We consider vertices with a distance no greater than a threshold $d_0$ as neighbours:
\begin{equation}
    \mathcal{E}_{vv}(v_{s},v_{t})= 1,\ \text{if}\ d(v_{s},v_{t})\leq d_0.
\end{equation}


The construction of our Human Gaussian Graph is illustrated in Algorithm~\ref{alg:construct}.

\subsection{Graph Operations} \label{intra}

Based on our Human Gaussian Graph, we introduce two types of operations for information aggregation and interaction. The two operations are illustrated in Figure~\ref{fig:pipeline} and Algorithm~\ref{alg:attn}.

\begin{algorithm}[t]
\caption{Graph Operations}\label{alg:attn}
\renewcommand{\algorithmicrequire}{\textbf{Input:}}
\renewcommand{\algorithmicensure}{\textbf{Output:}}
\begin{algorithmic}
\Require Human Gaussian Graph $\{\mathcal{V},\mathcal{G},\mathcal{E}_{vv},\mathcal{E}_{vg}\}$
\Ensure SMPL-aligned Gaussians $\mathcal{G}^{\text{smpl}}$
\For {$l \gets 1$ \textbf{to} $L$} \Comment stack L blocks
    \For {$n \gets 1$ \textbf{to} $N$} \Comment intra-node operation
        \State $B_n(\mathcal{G}) = \{g_m^t,\ \text{if}\ \mathcal{E}_{vg}(g_m^t,v_{n})=1\}$
        \State $q_n \gets\text{Attention}(q=q_n,k=v=B_n(\mathcal{G}))$
        \State $q_n \gets \Phi_{\text{FFN}}(q_n)$
    \EndFor
    \For {$V_n \in \{V_n\}^N_{n=1}$} \Comment inter-node operation
        \State $B_n(\mathcal{V}) = \{v_s,\ \text{if}\ \mathcal{E}_{vv}(v_{s},v_{n})=1\}$
        \State $q_n \gets\text{Attention}(q=q_n,k=v=B_n(\mathcal{V}))$
        \State $q_n \gets \Phi_{\text{FFN}}(q_n)$
    \EndFor
\EndFor

\For {$m \gets 1$ \textbf{to} $M$} 
    \State $V_n \gets V_{n}\text{ ,where } E_{vg}^{t_0} (m,n)=1\}$
    \State $g_{m}^{\text{smpl}} = \Phi_\text{Att}\left(q=g_{m}^{t_0},k=v=q_n\right) + g_{m}^{t_0}$
\EndFor

\end{algorithmic}
\end{algorithm}

\begin{table*}[h]
\centering
    \setlength{\abovecaptionskip}{0cm}
    \caption{
       Quantitative comparison on novel view synthesis and novel pose animation with other methods. ``Gen." indicates generalizable methods and ``Ani." denotes capability for novel pose animation. $\dagger$: LGM is fine-tuned on our dataset. As LGM lacks inherent animation capability, we directly bind its output to SMPL meshes for comprehensive comparative evaluation.
    }
    \label{tab:quantitative_a}
    \vspace{2mm}
    \resizebox{\textwidth}{!}{
    \renewcommand\arraystretch{1.3}

    \begin{tabular}{l|c|cc|ccc|ccc|c}
     \toprule[1.2pt]
      & \multicolumn{1}{c|}{\multirow{2}{*}{\textbf{Setting}}} & \multicolumn{2}{c|}{\textbf{Category}} & \multicolumn{3}{c|}{\textbf{Novel View}}& \multicolumn{3}{c|}{\textbf{Novel Pose}} & \multicolumn{1}{c}{\multirow{2}{*}{\textbf{Time}}} \\ 

    \multicolumn{1}{c|}{} &  & Gen. & Ani. & SSIM$\uparrow$ & PSNR$\uparrow$ & LPIPS$\downarrow$ & SSIM$\uparrow$ & PSNR$\uparrow$ & LPIPS$\downarrow$ 
      \\
    \hline 
    GART~\cite{lei2023gart} &Single & \no & \yes & 0.906 & 20.109 &0.098 & 0.866 & 18.964  & 0.124 & 1.9m \\
    ExAvatar~\cite{moon2024exavatar} &Single & \no & \yes & 0.910 & 21.450 & \textbf{0.077} & 0.894 & 19.327 & \textbf{0.087} &2.9h  \\
    LGM$^\dagger$~\cite{LGM:2024} &Single & \yes & \no & 0.912 & 21.274 & 0.093 & 0.889 & 19.560  &0.126  & \textbf{9.7s} \\

    \cellcolor{mygray}\textbf{Ours}  &\cellcolor{mygray}Single & \cellcolor{mygray}\yes & \cellcolor{mygray}\yes & \cellcolor{mygray}\textbf{0.920} & \cellcolor{mygray}\textbf{23.112} & \cellcolor{mygray}\underline{0.080} & \cellcolor{mygray}\textbf{0.896} & \cellcolor{mygray}\textbf{21.857} & \cellcolor{mygray}\underline{0.111} & \cellcolor{mygray}\textbf{9.7s} \\
    \hline
    4DGS~\cite{Wu_2024_CVPR} &Multi & \no & \no & 0.814 & 19.072 & 0.139 & - & - & - & 8.5m \\ 
    GPS-Gaussian~\cite{zheng2024gpsgaussian} &Multi & \yes & \no & 0.827 & 19.129 & 0.134 & - & - & - & \textbf{0.6s} \\

    \cellcolor{mygray}\textbf{Ours}  &\cellcolor{mygray}Multi & \cellcolor{mygray}\yes & \cellcolor{mygray}\yes & \cellcolor{mygray}\textbf{0.955} & \cellcolor{mygray}\textbf{26.536} & \cellcolor{mygray}\textbf{0.048} & \cellcolor{mygray}\textbf{0.934} & \cellcolor{mygray}\textbf{24.013} & \cellcolor{mygray}\textbf{0.066} & \cellcolor{mygray}\underline{0.9s} \\
    
     \bottomrule[1.2pt]
    \end{tabular}
}
    \vspace{-1em}
\end{table*}

\noindent\textbf{Intra-node Operation.}
To extract rich information of Gaussian from different timesteps, we propose an intra-node operation on $\mathcal{E}_{vg}$. 
For each mesh vertex $v_{n}$, we define a learnable query $q_n$ for capturing individual-agnostic features. 
We define the neighbor Gaussian group of a mesh vertex $v_{n}$ as:
\begin{equation}
    B_n(\mathcal{G}) = \{g_m^t,\ \text{if}\ \mathcal{E}_{vg}(g_m^t,v_{n})=1\}.
\end{equation}
Thus, the learnable query $q_n$ can be updated by the attention mechanism:
\begin{equation}
    \tilde{q}_{n} = \Phi_\text{Att}\left(q=q_n,k=v=B_n(\mathcal{G})\right),
\end{equation}
where the key, query and value of $\Phi_\text{Att}$ are formulated as:
\begin{align}
    Q& = h_{Q}(q_{n}), \\
    K& = h_{K}\left(B_n(\mathcal{G})\right), \\
    V& = h_{K}\left(B_n(\mathcal{G})\right), 
\end{align}
where $h_{Q}$, $h_{K}$, $h_{V}$ stand for projections.
After this, the output is passed through a standard feed-forward network in the transformer:
\begin{equation}
    q_{n} = \Phi_{\text{FFN}}(\tilde{q}_{n}).
\end{equation}

\noindent\textbf{Inter-node Operation} \label{inter}
To support message passing across mesh vertices, we further propose an inter-node operation on $\mathcal{E}_{vv}$. 
Similarly, we define the neighbor Gaussian group of a mesh vertex $v_{n}$ as:
\begin{equation}
    B_n(\mathcal{V}) = \{v_s,\ \text{if}\ \mathcal{E}_{vv}(v_{s},v_{n})=1\}.
\end{equation}
Thus, the learnable query $q_n$ can be updated by:
\begin{align}
    \tilde{q}_{n} = \Phi_\text{Att}\left(q=q_n,k=v=B_n(\mathcal{V})\right),
\end{align}
where $\Phi_\text{Att}$ stand for attention mechanism.
After this, the output is passed through a standard feed-forward network in the transformer:
\begin{equation}
    q_{n} = \Phi_{\text{FFN}}(\tilde{q}_{n}).
\end{equation}
The learnable query $\{q_n\}_{n=1}^{N}$ is aligned with the SMPL mesh, which has integrated features of Gaussians from different frames. 
Since the inter-node operation enables the model to achieve integration within a relatively local region, multiple operations can be stacked to broaden the receptive field of each node.

\subsection{Training Objectives} \label{smpl-aligned}

Given one chosen frame of initial Gaussian $G^{t_0}=\{g_{m}^{t_{0}}\}_{m=1}^{M}$, we introduce the aforementioned $\{q_n\}$ to refine the Gaussian features:
\begin{align}
    &g_{m}^{\text{smpl}} = \Phi_\text{Att}\left(q=g_{m}^{t_0},k=v=q_n\right) + g_{m}^{t_0}, \\
    &\text{where}\  n=\mathop{\arg\min}\limits_{i}\ \psi_{\text{d}}(\mu_{m}^{t_0}, v_i^t).
    \label{eq. smplattn}
\end{align}
The final SMPL Gaussians $G^{smpl}=\{g_{m}^{\text{smpl}}\}_{m=1}^{M}$ are temporal-invariant.
Specifically, for each frame with multi-view images $I^t_{mv}$, the Gaussian avatar is first re-posed into the estimated SMPL pose of the frame $p_t$.
Then, the re-posed Gaussians are rendered using the standard 3D Gaussian rasterization. 
Following the instruction of LGM~\cite{LGM:2024}, we use mean square error loss and LPIPS loss to the RGB image and mean square error loss on the alpha image:
\begin{equation}
    \mathcal{L} = \mathcal{L}_{\text{MSE}}(\tilde{I}_{\text{rgb}},I_{\text{rgb}})+\alpha_1\mathcal{L}_{\text{LPIPS}}(\tilde{I}_{\text{rgb}},I_{\text{rgb}})+\alpha_{2}\mathcal{L}_{\text{MSE}}(\tilde{I}_{\alpha},I_{\alpha}), \nonumber
\end{equation}
where $I_{\text{rgb}}$, $I_{\alpha}$ are the ground truth images and $\tilde{I}_{\text{rgb}}$, $\tilde{I}_{\alpha}$ are the corresponding rendered output images from our model.
\section{Experiment}
\begin{figure*}[]
\centering
\includegraphics[width=\linewidth]{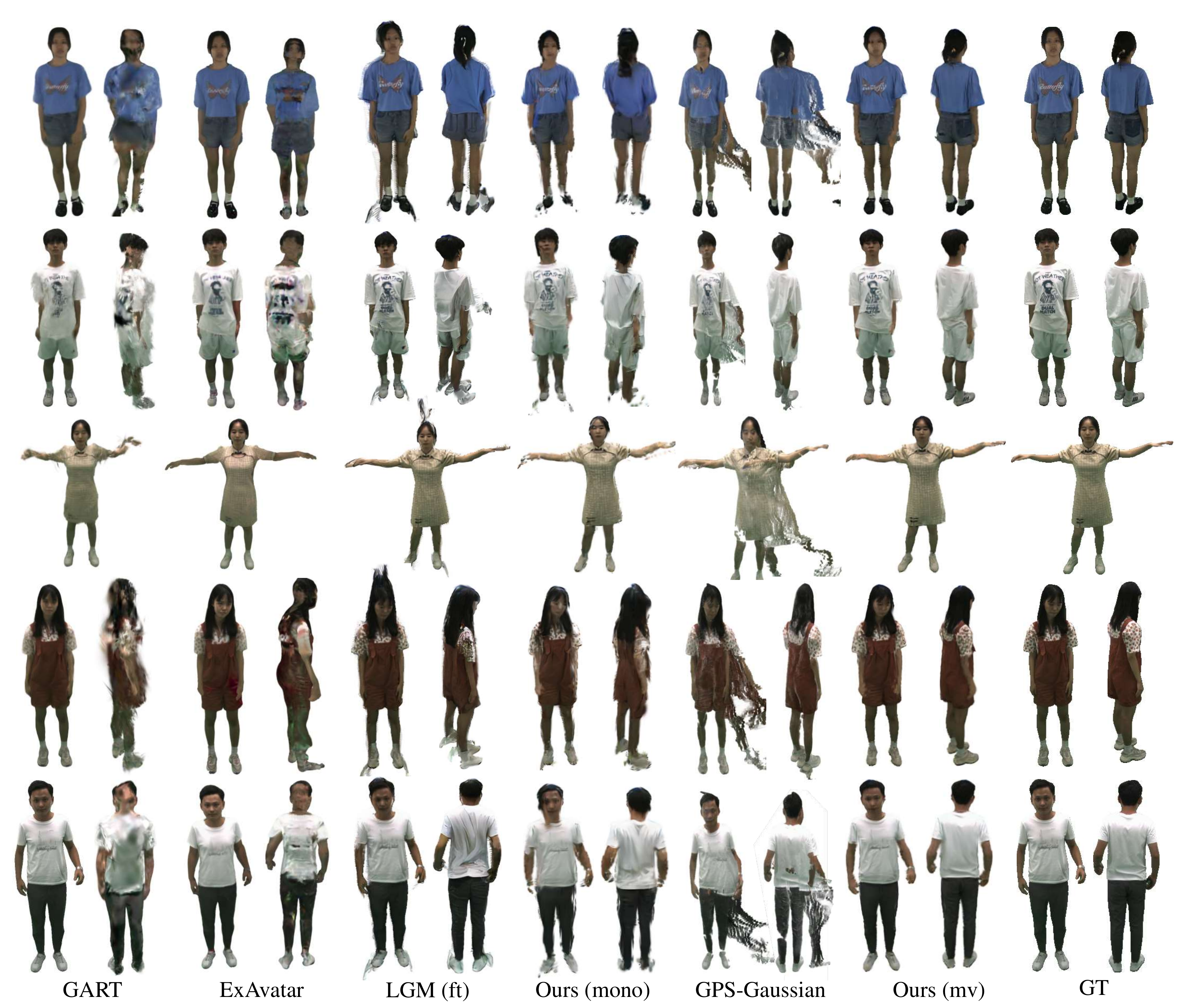}
\caption{Qualitative comparison of ours against GART~\cite{lei2023gart}, ExAvatar~\cite{moon2024exavatar}, LGM~\cite{LGM:2024} and GPS-Gaussian~\cite{zheng2024gpsgaussian} on MvHumanNet dataset~\cite{xiong2024mvhumannet}. Our approach achieves the highest visual fidelity and reconstruction quality in both single-view and multi-view setting. }
\label{fig:main results}
\end{figure*}

\subsection{Implementation Details} \label{sec:implementation}
\noindent\textbf{Dataset.}
We utilize MvHumanNet~\cite{xiong2024mvhumannet} as our training dataset. MvHumanNet is a large-scale multi-view video human dataset that includes estimated SMPL parameters. For our experiments, we curated a training split comprising 2,000 distinct individuals and sampled 60 frames from each subject's sequence. This sampling strategy yielded a diverse corpus of 120,000 poses, featuring varied individuals in a wide range of postures and movements. 

\noindent\textbf{Training.}
Following LGM~\cite{LGM:2024}, we leverage the four ground-truth views (front, left, back, and right) from the MvHumanNet dataset as input to the LGM U-Net. We render our posed SMPL-aligned Gaussians onto eight views, comprising the four input views plus four additional randomly sampled viewpoints. The loss function described in Section \ref{smpl-aligned} is computed between these rendered images and their corresponding ground-truth counterparts. We keep the LGM U-Net frozen throughout the training process, focusing optimization exclusively on the HGG modules. Training convergence is achieved within two days using eight NVIDIA A800 (40GB) GPUs.

\begin{figure*}[t]
\centering
\includegraphics[width=1.0\linewidth]{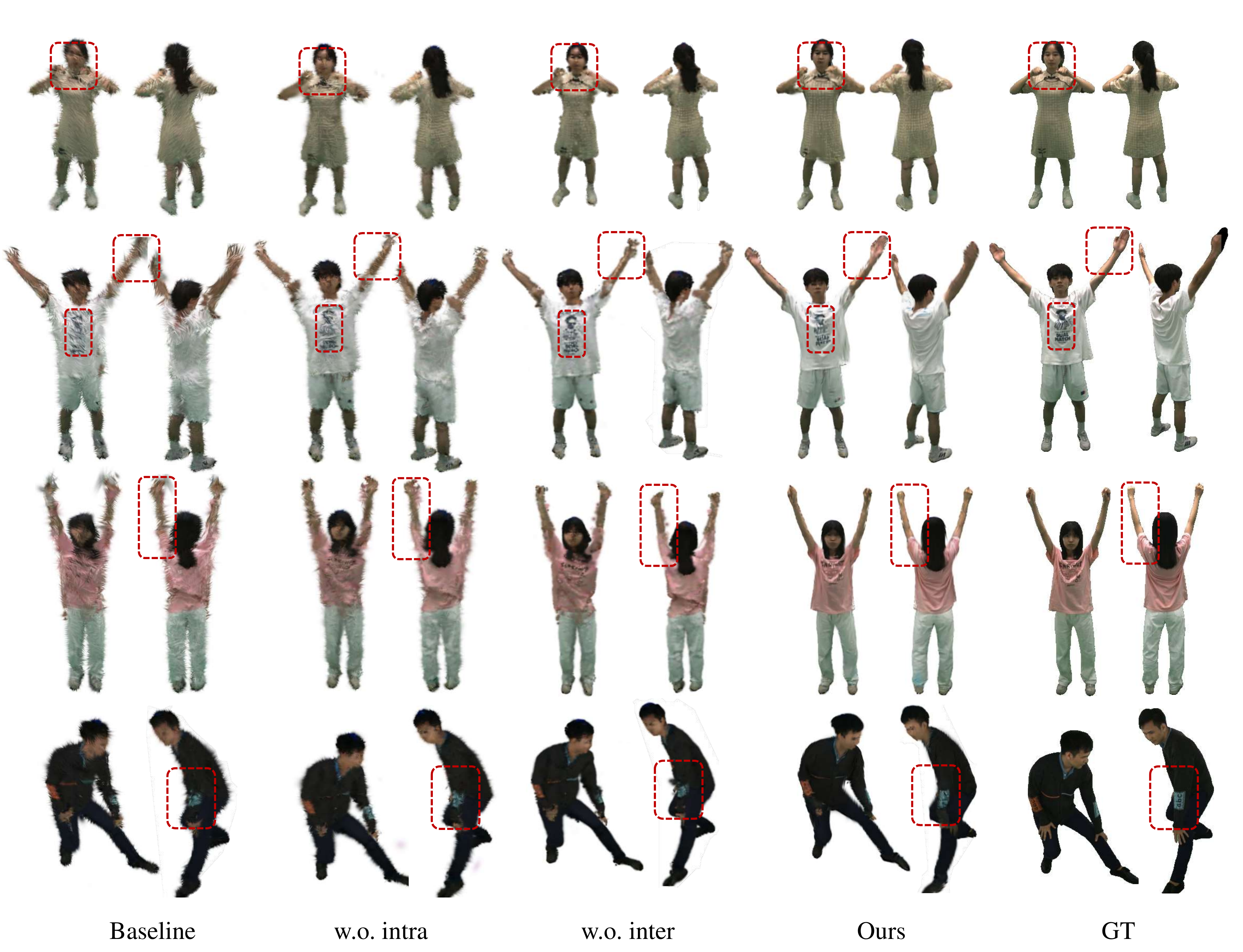}
\caption{Qualitative results for novel pose animation. We evaluated each module in our proposed Human Gaussian Graph and analyzed the growth. The results demonstrate the improvement with each component and the overall quality of our novel poses. We kindly refer readers to the supplementary for more visualization results.}
\vspace{-0.3cm}
\label{fig:pose}
\end{figure*}

\noindent\textbf{Inference time.} Our framework supports both multi-view and single-view video inputs. For multi-view processing, we directly feed four images per frame into the network, generating SMPL-aligned Gaussians in 0.9 seconds. For single-view input, we employ a fine-tuned Wonder3D~\cite{long2023wonder3d} multi-view diffusion model to synthesize multi-view videos, requiring 8.8 seconds for diffusion, resulting in a total inference time of 9.7 seconds per frame. Once constructed, the SMPL-aligned Gaussian Avatar renders novel views and poses at over 120 FPS on an A800 GPU.

\subsection{Comparison}
\noindent\textbf{Setting.} 
Given a video, the model should be able to output arbitrary views of the human in the video. Some of the methods could model an animatable avatar and generate views in novel poses. We evaluated the quality of the novel view and pose using PSNR, SSIM~\cite{SSIM:2004} and LPIPS~\cite{zhang2018perceptual} metrics.
Table~\ref{tab:quantitative_a} presents a comprehensive comparison between our approach and contemporary Gaussian methods across both monocular and multi-view video settings. Gaussian methods fall into two categories: optimization-based approaches such as 4DGS~\cite{wu20234d}, GART~\cite{lei2023gart}, and ExAvatar~\cite{moon2024exavatar}; and generalizable methods including LGM~\cite{LGM:2024} and GPS-Gaussian~\cite{zheng2024gpsgaussian}.

We conduct a quantitative comparison on 10 MvHumanNet~\cite{xiong2024mvhumannet} scans, with detailed information about the evaluation split available in the supplementary material. To ensure fairness, we fine-tuned LGM~\cite{LGM:2024} on our training dataset and fully optimized each optimization-based method by selecting their longest training steps.

\begin{table*}[htbp]
    \caption{Ablation studies. (a) ``Baseline" stands for the model without Human Gaussian Graph and learnable mesh-aligned queries, ``intra" represents the intra-node transformer and ``inter" is the inter-node transformer. Each component can bring improvement to the architecture. (b) While 1 stack of ``intra inter" pair already boosts the performance, adding more layers would be beneficial. Metrics are evaluated on the novel pose task in the multi-view setting.}
    \label{fig:ablation_studies}

    \vspace{-0.2cm}
    \begin{subtable}[t]{0.5\textwidth}
        \centering
        \renewcommand\arraystretch{1.3}
        \subcaption{Ablation study for proposed modules}
        \begin{tabular}{l|ccc}
            \toprule[1.2pt]
            \textbf{Method} & SSIM$\uparrow$ & PSNR$\uparrow$ & LPIPS$\downarrow$ \\
            \hline
            Baseline             & 0.915\textsubscript{\textcolor{gray}{(-0.019)}} & 21.327\textsubscript{\textcolor{gray}{(-2.686)}} & 0.092\textsubscript{\textcolor{gray}{(+0.026)}}   \\
            w.o. intra     & 0.930\textsubscript{\textcolor{gray}{(-0.004)}} & 21.902\textsubscript{\textcolor{gray}{(-2.111)}} & 0.090\textsubscript{\textcolor{gray}{(+0.024)}}  \\
            w.o. inter            & 0.931\textsubscript{\textcolor{gray}{(-0.003)}} & 23.105\textsubscript{\textcolor{gray}{(-0.908)}} & 0.087\textsubscript{\textcolor{gray}{(+0.021)}}  \\
            \hline
            \cellcolor{mygray}\textbf{Ours}   & \cellcolor{mygray}\textbf{0.934} & \cellcolor{mygray}\textbf{24.013} & \cellcolor{mygray}\textbf{0.066}  \\
            \bottomrule[1.2pt]
        \end{tabular}
        
        \label{tab:ablation_blocks}
    \end{subtable}
    \hfill
    \begin{subtable}[t]{0.5\textwidth}
        \centering
        \renewcommand\arraystretch{1.3}
        \subcaption{Ablation study for scaling up}
        \begin{tabular}{l|ccc}
            \toprule[1.2pt]
            \multicolumn{1}{c|}{\textbf{Num Layers}} & SSIM$\uparrow$ & PSNR$\uparrow$ & LPIPS$\downarrow$ \\ 
            \hline
            L = 0         &   0.930\textsubscript{\textcolor{gray}{(-0.004)}}  & 21.902\textsubscript{\textcolor{gray}{(-2.111)}}  &  0.090\textsubscript{\textcolor{gray}{(+0.024)}} \\
            L = 1         &   0.926\textsubscript{\textcolor{gray}{(-0.008)}}   &   23.455\textsubscript{\textcolor{gray}{(-0.558)}}   &   0.071\textsubscript{\textcolor{gray}{(+0.005)}}  \\
            L = 3         &   0.929\textsubscript{\textcolor{gray}{(-0.005)}}   &   23.925\textsubscript{\textcolor{gray}{(-0.088)}}   &    0.073\textsubscript{\textcolor{gray}{(+0.007)}}  \\ 
            \hline
            \cellcolor{mygray}\textbf{Ours} (L = 6)      &  \cellcolor{mygray}\textbf{0.934} & \cellcolor{mygray}\textbf{24.013} & \cellcolor{mygray}\textbf{0.066}  \\
            \bottomrule[1.2pt]
        \end{tabular}
        
        \label{tab:ablation_L}
    \end{subtable}

\vspace{-0.1cm}
\end{table*}

\noindent\textbf{Quantitative Comparison.}  As shown in Table~\ref{tab:quantitative_a}, our method outperforms the counterparts on both single-view setting and multi-view setting, as well as both novel view synthesis task and novel pose animation task, demonstrating the effectiveness of our design. Specifically, on novel view synthesis for single view, we achieve a boost of 1.6 dB of PSNR against ExAvatar~\cite{moon2024exavatar}. Notably, ExAvatar excels in front view reconstruction and especially face modeling, but our overall reconstruction results are better than theirs, indicating the strong 3D reconstruction capabilities of our model thanks to the Human Gaussian Graph. Our advantage over LGM~\cite{LGM:2024}, which takes one frame as input, demonstrate the value of gathering information across the frames. For multi-view reconstruction, ours method is significantly ahead. Current methods like 4DGS~\cite{Wu_2024_CVPR} and GPS-Gaussian~\cite{zheng2024gpsgaussian} are designed only for rather dense-view inputs (i.e. 16 views for GPS-Gaussian), and perform sub-optimally when the input multi-view is sparse. In contrast, our method showcase efficient usage of the multi-view information via HGG, enabling realistic reconstruction results. For novel pose animation, our method also demonstrates a significant advantage of 2.3 dB of PSNR. This indicates the high-quality of our reconstructed animatable avatar represented by SMPL-aligned Gaussians, fueling a variety of downstream applications and tasks.

\noindent\textbf{Qualitative Comparison.}
As illustrated in Figure~\ref{fig:main results}, by integrating cross-frame information via HGG, our model achieves higher fidelity in novel view synthesis against counterparts. For single-view setting, optimization-based methods like GART~\cite{lei2023gart} and ExAvatar~\cite{moon2024exavatar} reconstructs high-quality training views, but neglects 3D consistency and result in collapsed novel views. LGM~\cite{LGM:2024} can only leverage one frame, and suffers from artifacts generated by the multi-view diffusion in its pipeline. Though our method also faces the multi-view diffusion problem, HGG can effectively suppress the artifacts with extra knowledge from other frames and neighboring second-layer nodes. For multi-view setting, GPS-Gaussian~\cite{zheng2024gpsgaussian} suffers severe hallucination when fusing the sparse multi-views. Our method achieves consistently photorealistic quality with clear contour and satisfactory details, demonstrating the effectiveness of HGG. Notably, our method achieves high-quality face and hand reconstruction without bells and whistles, which is regarded as a severe obstacle~\cite{pan2024humansplat, moon2024exavatar}.


\subsection{Ablation Study}

We conduct an ablation study on the challenging novel pose animation for each of the proposed modules in our HGG. The baseline model is to directly bind predicted Gaussians to SMPL and drive them with pose parameters. We provide quantitative results in~\ref{tab:ablation_blocks} and qualitative results in Fig.~\ref{fig:main results}.

\noindent\textbf{Learnable mesh queries.} The learnable mesh queries, binding to the SMPL vertices, serve as information collectors and human-agnostic bias. Even without the graph architecture and the information-gathering process, the learnable queries can still provide a slight enhancement (i.e. 0.57 in PSNR). We attribute this to the basic human obtained via massive training, which will help regulate the Gaussian-aligning process, providing a boost in visual qualities.

\noindent\textbf{Intra-node operation.}
The intra-node operation is aimed for efficiently extracting information from the Gaussian nodes across the temporal axis. Comparing Line 2 and 4 in Table~\ref{tab:ablation_blocks}, the intra-node operation brings about a boost for 2.1 dB in PSNR. As shown in Figure~\ref{fig:pose}, the intra-node operation significantly promotes the details such as in the face, the hands, and the printed patterns on the clothes. It demonstrates that information from other frames can indeed enhance the quality of Gaussians, underlying the importance of cross-frame interaction. 

\noindent\textbf{Inter-node operation.}
The inter-node operation, enabling neigbour communication across second-layer nodes, is effective when intra-node operation is active. Comparing Line 3 and 5 in Table~\ref{tab:ablation_blocks}, adding inter-node operation can bring about a further 0.9 dB of PSNR boost. This result proves the necessity of feature aggregation and purification in localities in the second layer. Effectiveness of this module is also illustrated in Fig. \ref{fig:pose}, where the noise and aliases are significantly suppressed.

\noindent\textbf{Scaling up.}
Theoretically, stacking the "inter-inter" pairs in Section~\ref{inter} can enlarge the perception field for each second-layer node, analogous to convolutions. We conducted an experiment with the number of layers stacked in Table~\ref{tab:ablation_L}, which shows the increasing number of layers can demonstrably boost performance.

\section{Conclusion} \label{conclusion}
We present a pioneering generalizable and animatable Gaussian human network that derives SMPL-aligned Gaussians from monocular or multi-view videos within inference time. This model proposes a novel dual-level Human Gaussian Graph, enabling effective and efficient feature communication both across the temporal axis and within the topological neighbourhood. Extensive quantitative experiments demonstrate that our method surpasses existing methods in both single-view and multi-view settings, including novel view synthesis and novel pose animation. This capability opens up a broad spectrum of downstream applications.

\noindent\textbf{Limitations and Future works.} We observed some gaps in the experiment results that could potentially be addressed by future researches. 
(1) Quality gaps between monocular setting and multi-view setting. The monocular setting lags behind multi-view results in a large scale (i.e. 3.4 dB in PSNR). We attribute it to the absence of open source realistic human diffusion model. 
Researches on human diffusion models would fuel this field. 
(2) Detail degradation for novel poses. Though we achieve SOTA in novel pose animation, there is a gap to the results in novel views (i.e. 1.5-2.5 dB in PSNR). We attribute it to the inaccurate SMPL parameters from the dataset. Future works could address this by introducing SMPL parameter refinement techniques.

\noindent\textbf{Acknowledgments.} This work was supported in part by the National Natural Science Foundation of China under Grant 62206147, and in part by 2024 WeChat Vision, Tecent Inc. Rhino-Bird Focused Research Program.

{
    \small
    \bibliographystyle{ieeenat_fullname}
    \bibliography{main}
}

\clearpage
\setcounter{page}{1}
\maketitlesupplementary

\section{Implementation Details}
\noindent\textbf{Fine-tuning multi-view diffusion.}
We utilize the Wonder3D~\cite{long2023wonder3d} as the multi-view diffusion model. As the Wonder3D model is designed for objects and trained on Objaverse~\cite{deitke2023objaverse}, it has limited knowledge about humans and directly applying it into our architecture would lead to dissatisfactory results. To mitigate this problem, we fine-tuned Wonder3D with the MvHumanNet dataset. To ensure training efficiency, we only selected 600 humans and used 30 poses for each scan, summing to 18K training data. We input the front view and supervise the output with corresponding front, back, left and right views. The resolution is set to 256, and we use a learning rate of 5e-5 at the mixed training stage and 2.5e-5 at joint training stage. The training converges on 8 $\times$ Nvidia A800 GPUs in 9 days, with a batch size of 4 per GPU. Though improved, the overall result is still non-optimal, leading to the gap between the monocular setting and the multiview setting in our experiments.

\noindent\textbf{Fine-tuning Gaussian reconstruction model.} To obtain an initial set of Gaussians mentioned in Section 3.2:
\begin{equation}
    G^{t}=\{g_m^t\}_{m=1}^M, \quad g_m^t=\{\mu_{m}^t, f_{m}^t\},
\end{equation}
we leverage a fine-tuned LGM model. We use the same dataset mentioned in Section 4.1, and fine-tuned the ``large"" LGM with an input resolution of 256 and an output Gaussian resolution of 128 per view. We follow the original LGM configuration for the learning rate and batch size. The training converges on 8 $\times$ Nvidia A800 GPUs in 5 days.

\noindent\textbf{Training details.}
We train our HGG modules with the fine-tuned LGM model frozen. We uniformly sample 8 frames from each video and use the 8 frames as input for our module. The learning rate is set to 4e-4, gradient clip is 1.0, batch size to 1 per GPU and gradient accumulation steps to 8. We trained our model on 8 $\times$ Nvidia A800 GPUs, and it converges in 18 hours.

\noindent\textbf{Evaluation split.}
Our evaluation split is separated from the training split. We randomly selected 10 scans in the MvHumanNet as the evaluation split. Their IDs are listed as follows: 200102, 200114, 200125, 200134, 200137, 200151, 200535, 202148, 202209, 204157. 

\section{More Analysis}
\subsection{Efficiency}
Though processing large amount of information across the frames, HGG is highly-efficient thanks to the design of learnable queries, negligible compared with the LGM U-net. In this paragraph, the efficiency of each module will be theoretically analyzed. 

The intra-node transformer enables efficient communication with the Gaussians. If attention is directly applied to the union of all Gaussian sets, the complexity would be $O(M^2T^2D^2)$, M is the number of feed-forward Gaussians per frame, T is the number of the frames, and D denotes the dimension of the features. Given that N is often at the 10,000 level and a video typically have hundreds of frames, it would become an unacceptable computation bottleneck both for time and memory. With our HGG, Gaussians are collected by the mesh vertices, and the complexity becomes linear when each mesh node applies cross attention with the affiliated Gaussians with learnable queries: 
\begin{equation}
    O((MT/N)*N*D^2) = O(MTD^2)
\end{equation}
This would be over $10^6$ times more efficient than vanilla cross attention between Gaussians.

On the other hand, the computation complexity of the inter-node attention is rather small, thanks to the connectivity given by mesh. Empirically, a node has approximately 10 neigbours, so the complexity is only $O(10ND^2)$, which is negligible compared with other modules.

\begin{figure}[]
\centering
\includegraphics[width=\linewidth]{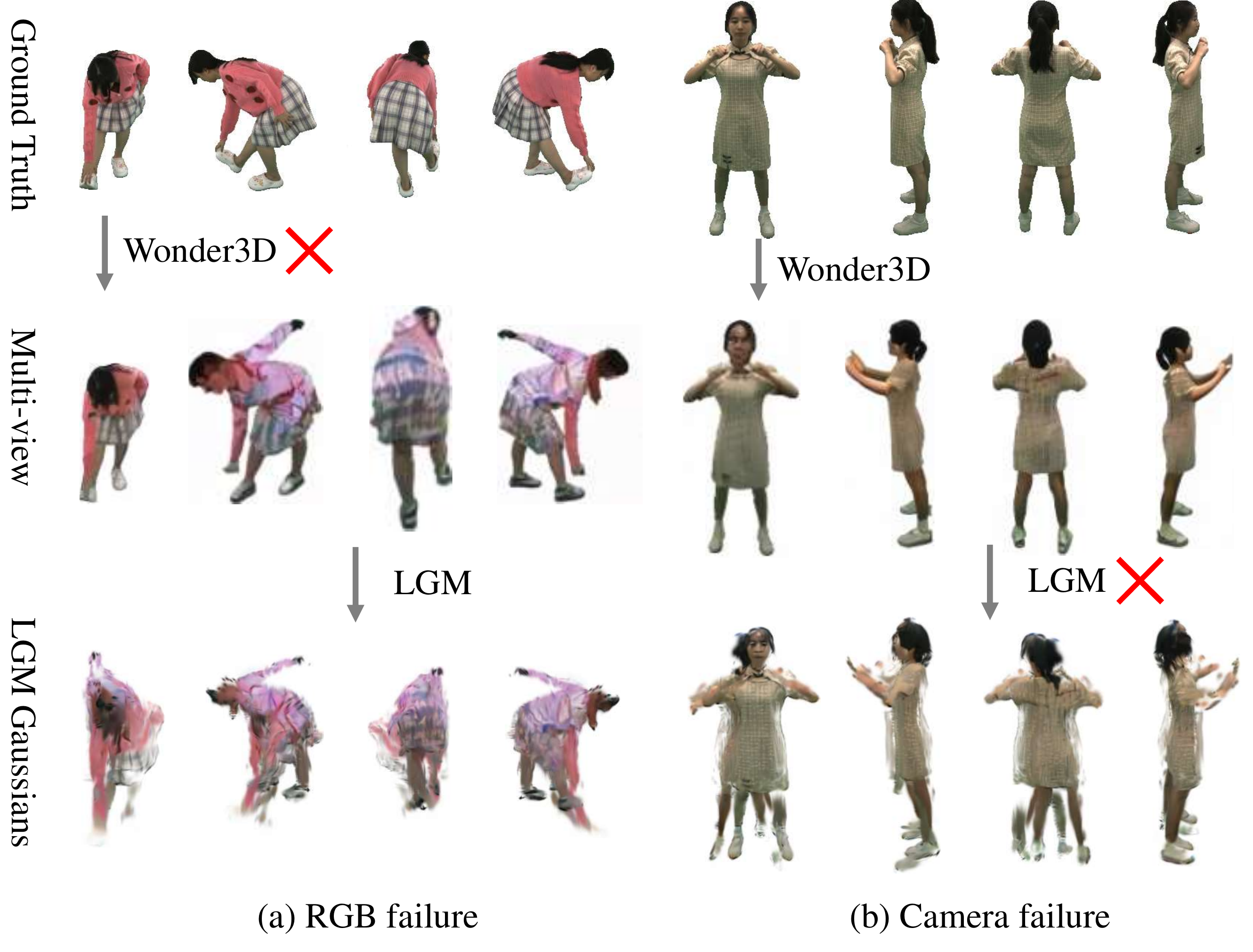}
\caption{Fail cases. (a) Wonder3D fails to generate reasonable back and side views, resulting in the failure for LGM Gaussian reconstruction. (b) Wonder3D generates multi-views of good quality, but LGM reconstruction fails due to inconsistent camera constraints provided by Wonder3D.}
\label{fig:fail}
\end{figure}

\begin{figure*}[ht]
\centering
\includegraphics[width=0.9\linewidth]{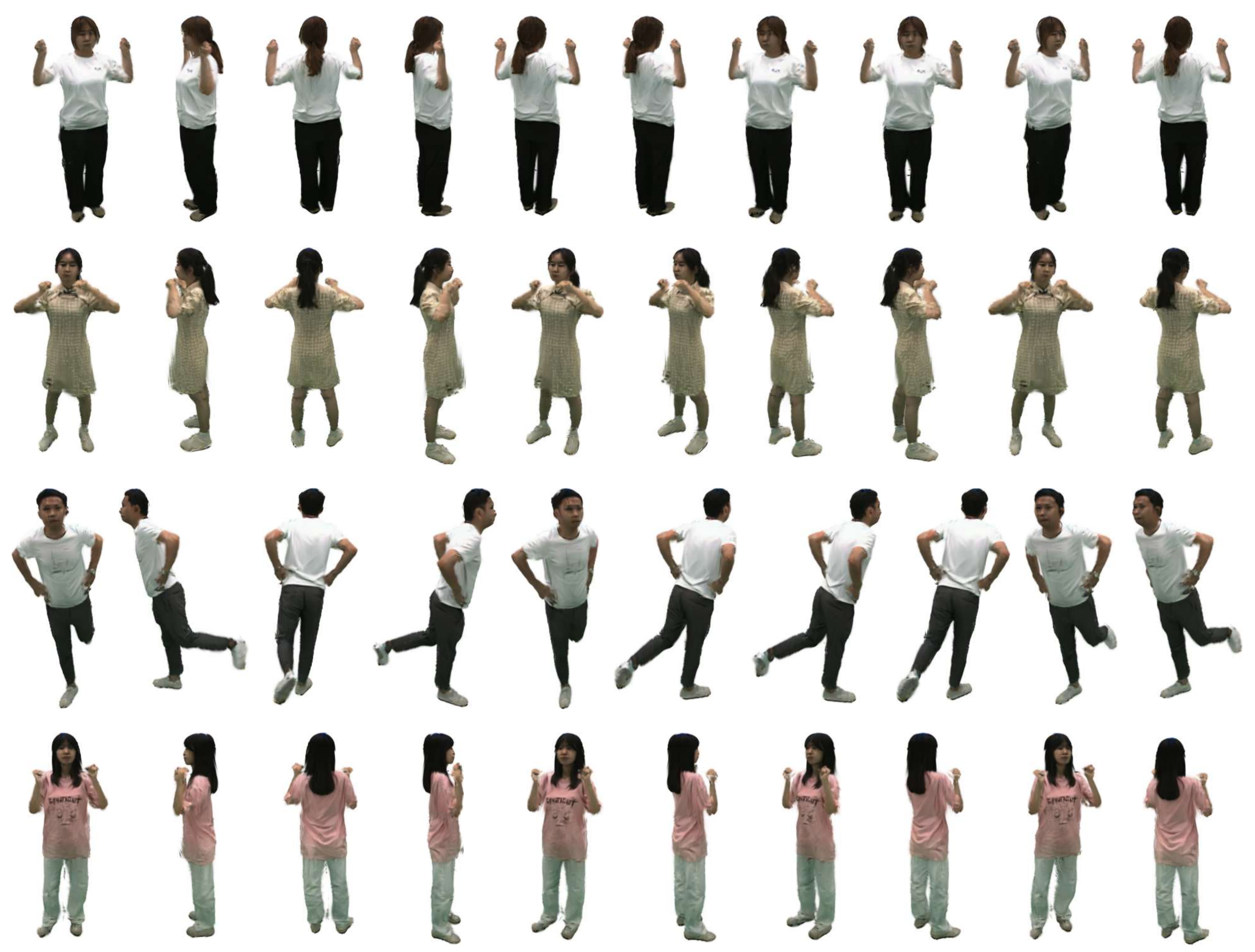}
\caption{More qualitative results on novel pose animation. The human avatar in various poses indicates the high-quality of our reconstructed 3D avatar.}
\label{fig:more_pose}
\end{figure*}

\subsection{Monocular Setting}
As introduced in the limitations, we observe a gap of 3.4 dB in PSNR between monocular settings and multi-view settings. Though we achieved SOTA in monocular setting, the quality is still far from downstream applications. As illustrated in Figure \ref{fig:fail}, we found that the fail cases largely derives from (1) the total failure of Wonder3D to generate novel views. (2) generated images do not follow camera constraint strictly, causing misalignments across views. These two reasons account for the failure in Gaussian initialization with LGM, and thus lead to corrupted results.

We attribute this issue to a lack of open-sourced real-world human diffusion models. Such work will largely fuel the field of single-view human reconstruction.

\begin{table}[t]
\centering
\vspace{-0.8em}
    \setlength{\abovecaptionskip}{0cm}
    \caption{
       Quantitative results of 50 testing examples.
    }
    \vspace{-0.5em}
    \tiny
    \label{tab:quantitative_rebuttal}
    \vspace{2mm}
    \resizebox{0.48\textwidth}{!}{
    \renewcommand\arraystretch{1.0}
    \setlength{\tabcolsep}{3pt}

    \begin{tabular}{l|ccc|ccc}
     \toprule[0.7pt]
      & \multicolumn{3}{c|}{\textbf{Deepfashion}} & \multicolumn{3}{c}{\textbf{MvHumanNet}} \\ 
      \multicolumn{1}{c|}{} & PSNR$\uparrow$ & SSIM$\uparrow$ & LPIPS$\downarrow$ & PSNR$\uparrow$ & SSIM$\uparrow$ & LPIPS$\downarrow$ \\
    \hline 
    LGM & 18.01 &   0.846 & 0.196 & 19.54 & 0.887& 0.129 \\
    \cellcolor{mygray}Ours (LGM) & \cellcolor{mygray}20.31
    & \cellcolor{mygray}0.901
    & \cellcolor{mygray}0.172
    & \cellcolor{mygray}21.82
    &\cellcolor{mygray} 0.892
    & \cellcolor{mygray}0.112 \\
    IDOL & 20.24  & 0.904 & 0.174 & 21.03 & 0.894 & 0.116 \\
    \cellcolor{mygray}{Ours (IDOL)} 
    & \cellcolor{mygray}\textbf{22.38}
    & \cellcolor{mygray}\textbf{0.912}
    & \cellcolor{mygray}\textbf{0.154}
    & \cellcolor{mygray}\textbf{23.59}
    & \cellcolor{mygray}\textbf{0.930}
    & \cellcolor{mygray}\textbf{0.092} \\
    \bottomrule[0.7pt]
    \end{tabular}
}
    \vspace{-1.8em}
\end{table}

\subsection{Comparison with new methods} 
IDOL can replace LGM to serve as the single-frame reconstruction module in our pipeline. Therefore, we further evaluate our method with this module. As shown in Table~\ref{tab:quantitative_rebuttal}, our method surpasses the SOTA method and achieves better results with the stronger backbone IDOL. 
Yet, AniGS has not open-sourced their codes or test splits.

\begin{figure*}[t]
  \centering
  \includegraphics[width=\linewidth]{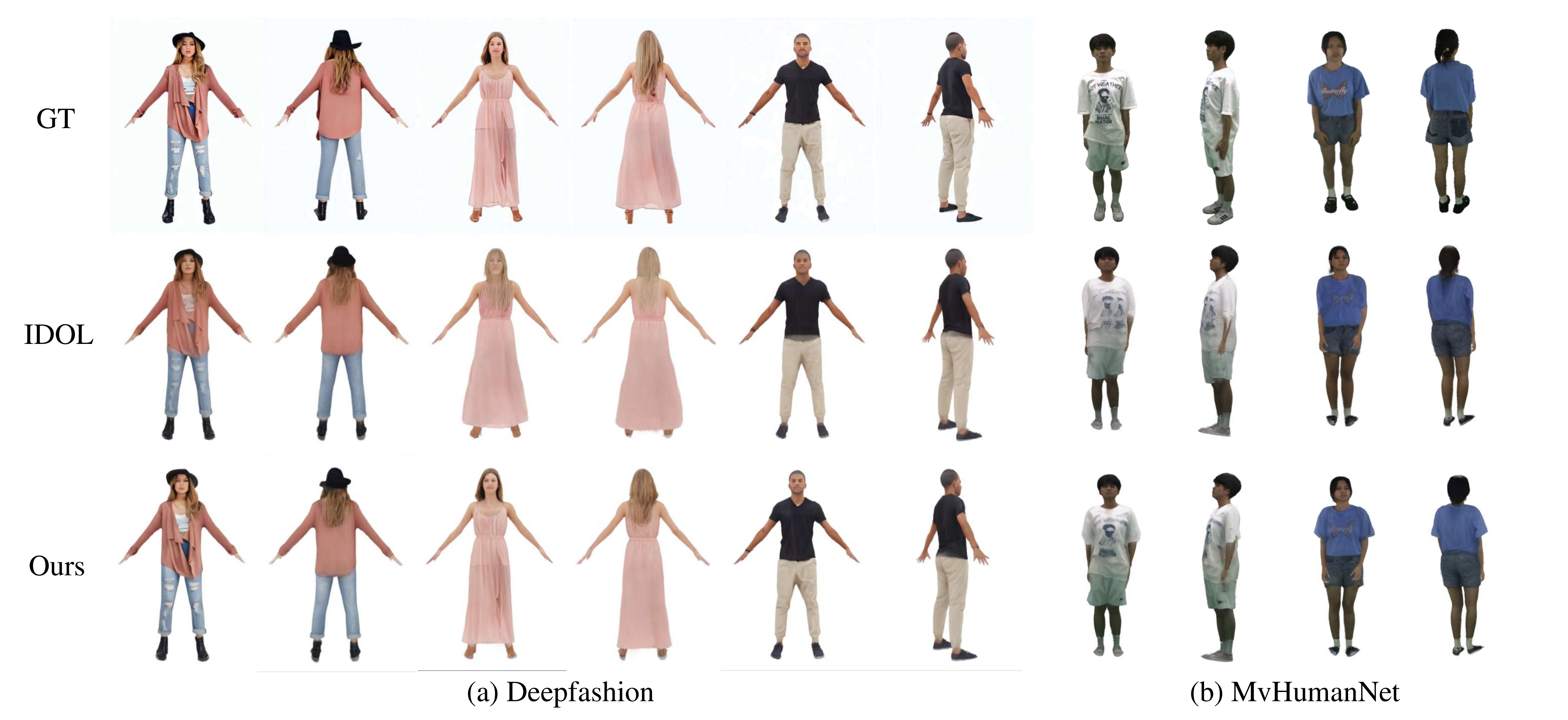}
  \vspace{-2.0em}
   \caption{Visualization on Deepfashion and MVHumanNet.}
   \label{fig:idol}
\end{figure*}

\begin{figure*}[ht]
\centering
\includegraphics[width=0.9\linewidth]{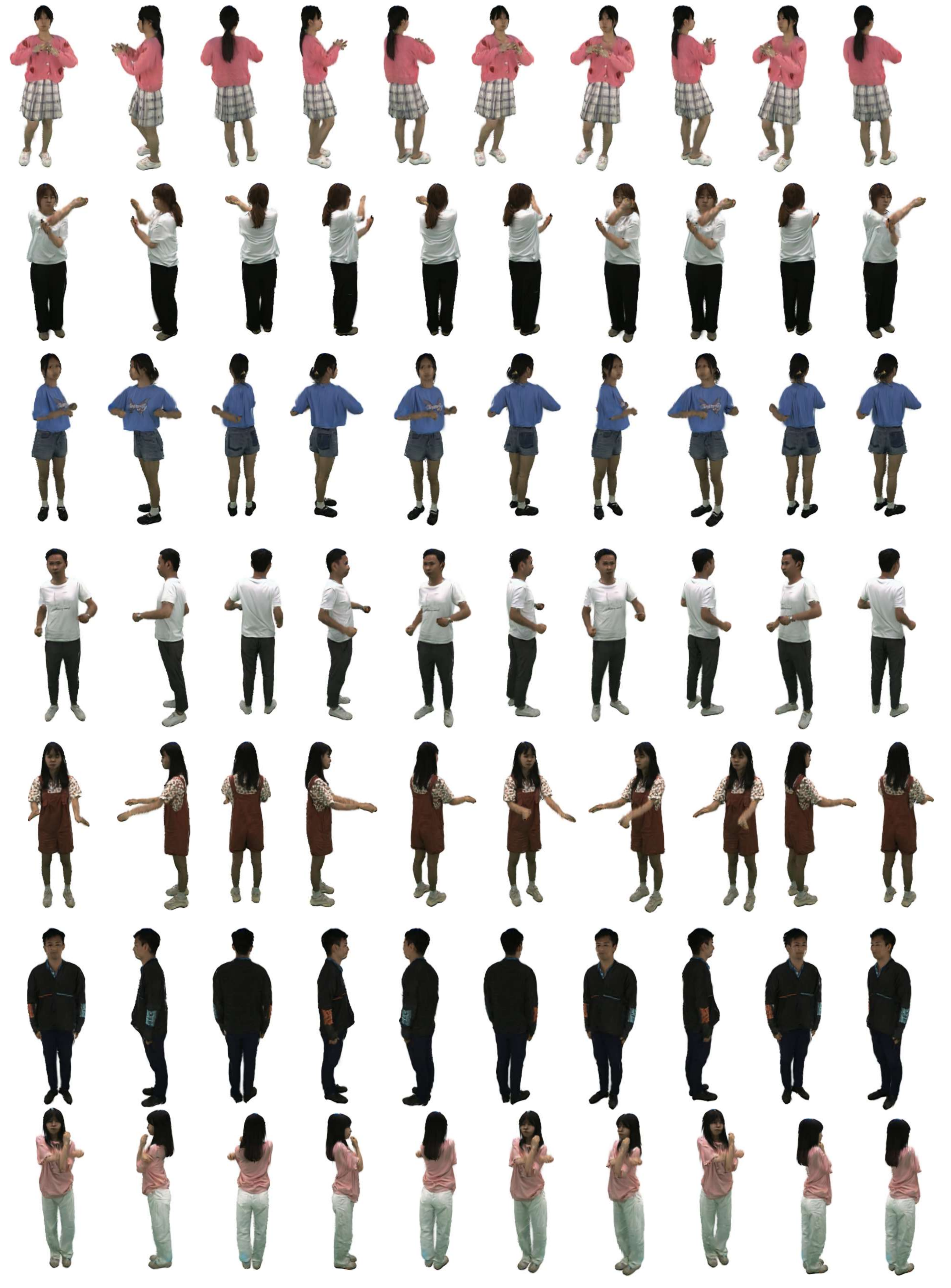}
\caption{More qualitative results on novel view synthesis. The novel views in multiple directions indicate the high-quality and potential downstream applications of our reconstructed 3D avatar.}
\label{fig:more_viz}
\end{figure*}

We conduct experiments on Deepfashion, an in-the-wild fashion clothing dataset. As shown in Table~\ref{tab:quantitative_rebuttal} and Figure~\ref{fig:idol}, our model achieves consistent performance improvements with different reconstruction modules.

\section{More Visualization Results}
We present more visualization results for both novel view synthesis and novel pose animation in Figure \ref{fig:more_pose} and \ref{fig:more_viz}.

\section{Broader Impacts}
Our model's capacity to generate high-quality 3D animatable avatars raises substantial privacy risks. To address these, the creation of ethical guidelines and legal frameworks is imperative. This necessitates close collaboration among researchers, developers, and policymakers.
Researchers should embed ethical considerations in development, while developers must implement privacy-centric practices. Policymakers need to craft regulations that define proper use, penalize misuse, and safeguard user privacy. Such collaboration is crucial for promoting the responsible application of this technology.


\end{document}